\documentclass{article}

\usepackage[final]{neurips_2020}
\usepackage[utf8]{inputenc}
\usepackage[T1]{fontenc}
\usepackage{hyperref}
\usepackage{url}
\usepackage{booktabs}
\usepackage{amsfonts}
\usepackage{nicefrac}
\usepackage{microtype}
\usepackage{algorithm}
\usepackage{algorithmic}
\usepackage{bbm}
\usepackage{amsmath}
\usepackage{graphicx}
\usepackage{makecell}
\usepackage{natbib}
\usepackage{wrapfig}

\DeclareMathOperator*{\argmin}{arg\,min}

\newcolumntype{?}{!{\vrule width 1pt}}
\newcolumntype{/}{!{\vrule width 0.7pt}}

\title{ZORB: A Derivative-Free Backpropagation Algorithm for Neural Networks}

\author{%
  Varun Ranganathan \\
  Department of Computing Science \\
  University of Alberta \\
  Edmonton, AB T6G 2E8 \\
  \texttt{vrangana@ualberta.ca} \\
  \And
  Alex Lewandowski \\
  Department of Computing Science \\
  University of Alberta \\
  Edmonton, AB T6G 2E8 \\
  \texttt{lewandowski@ualberta.ca} \\
}

\begin{document}

\maketitle

\begin{abstract}
Gradient descent and backpropagation have enabled neural networks to achieve remarkable results in many real-world applications.
Despite ongoing success, training a neural network with gradient descent can be a slow and strenuous affair.
We present a simple yet faster training algorithm called Zeroth-Order Relaxed Backpropagation (ZORB).
Instead of calculating gradients, ZORB uses the pseudoinverse of targets to backpropagate information.
ZORB is designed to reduce the time required to train deep neural networks without penalizing performance.
To illustrate the speed up, we trained a feed-forward neural network with 11 layers on MNIST and observed that ZORB converged 300 times faster than Adam while achieving a comparable error rate, without any hyperparameter tuning.
We also broaden the scope of ZORB to convolutional neural networks, and apply it to subsamples of the CIFAR-10 dataset. 
Experiments on standard classification and regression benchmarks demonstrate ZORB's advantage over traditional backpropagation with Gradient Descent.
\end{abstract}

\section{Introduction}

The workhorse of the deep learning paradigm is the Back-Propagation (BP) algorithm \citep{rumelhart1986learning}, which combines Gradient Descent (GD) with the chain-rule and caching \citep{ruder2016overview}.
Despite the ongoing success of Neural Networks (NNs), there are several theoretical limitations of training NNs with the GD-based BP algorithm.

\textbf{Slow to converge \citep{gaivoronski1994convergence}:}
GD calculates errors based on a first-order approximation of the error function, which becomes more inaccurate as we move away from the initial values of the parameters.
This requires the recalculation of gradients after every update operation, creating a bottleneck in training time.
ZORB updates the parameters of the network only once, which results in faster convergence.

\textbf{Hyperparameters:}
Several hyperparameters, such as learning rate, momentum \citep{qian1999momentum}, nestorov \citep{nesterov1983method}, batch size and epochs, need to be tuned for successful GD-based training of NNs \citep{guo2001pseudoinverse}.
Optimal hyperparameter values vary widely across datasets, requiring the practitioner to perform hyperparameter tuning \citep{bergstra2012random}, which trains multiple models at the expense of compute power.
This prevents machines with low compute power and real-time systems from using deep neural networks \citep{nishihara2017real}.
ZORB does not require hyperparameters, thereby reducing the time required to develop production-grade models.

We propose ZORB, a derivative-free BP algorithm, to reduce the training time of a NN without penalizing performance.
ZORB utilizes the Moore-Penrose pseudoinverse \citep{golan2012moore} to propagate information and update weights in the backward pass.
To counter the problem of invertibility, we also introduce adaptive activation functions that perform linear operations on the input and output matrices.

\begin{figure}[h]
\begin{minipage}[]{0.44\linewidth}
\centering
\resizebox{\columnwidth}{!}{
\begin{tabular}{|c|c?c|c/c|}
    \hline
    \textbf{\#L} & \multicolumn{1}{c?}{\textbf{Metric}} & \textbf{Adam} & \textbf{MLELM} & \textbf{ZORB} \\ \hline
    \Xhline{2.5\arrayrulewidth}
     & \textbf{Train Err.} & $0.090$ & $0.884$  & $\textbf{0.079}$  \\ \cline{2-5}
     &  \textbf{Train Acc.} & $\textbf{95.29}$ & $82.26$ & $94.43$ \\ \cline{2-5}
    \textbf{6} & \textbf{Test Err.} & $\textbf{0.095}$ & $0.883$  & $0.105$  \\ \cline{2-5}
     &  \textbf{Test Acc.} & $\textbf{94.83}$ & $82.27$ & $92.52$ \\ \cline{2-5}
     & \textbf{Time (s)} &  $5.8\small{\times}10^4$  & $178.815$ & $\textbf{135.125}$ \\
     \Xhline{2\arrayrulewidth}
     & \textbf{Train Err.} & $0.097$ & $0.899$  & $\textbf{0.079}$  \\ \cline{2-5}
     &  \textbf{Train Acc.} & $\textbf{95.19}$ & $73.55$ & $94.47$ \\ \cline{2-5}
    \textbf{8} & \textbf{Test Err.} & $\textbf{0.098}$ & $0.898$  & $0.113$  \\ \cline{2-5}
     & \textbf{Test Acc.} & $\textbf{94.96}$ & $74.21$ & $91.93$ \\ \cline{2-5}
     & \textbf{Time (s)} & $7.9\small{\times}10^4$ & $288.693$ & $\textbf{207.276}$ \\
     \Xhline{2\arrayrulewidth}
     & \textbf{Train Err.} & $0.117$ & $0.925$  & $\textbf{0.074}$  \\ \cline{2-5}
     & \textbf{Train Acc.} & $93.90$ & $52.37$ & $\textbf{94.79}$ \\ \cline{2-5}
    \textbf{11} & \textbf{Test Err.} & $0.120$ & $0.925$  & $\textbf{0.111}$  \\ \cline{2-5}
     & \textbf{Test Acc.} & $\textbf{93.61}$ & $53.11$ & $92.15$ \\ \cline{2-5}
     & \textbf{Time (s)} & $1.1\small{\times}10^5$ & $543.330$ & $\textbf{376.133}$ \\
     \Xhline{2\arrayrulewidth}
\end{tabular}
}
\end{minipage}
\begin{minipage}[]{0.55\linewidth}
\centering
\resizebox{\columnwidth}{!}{
\includegraphics{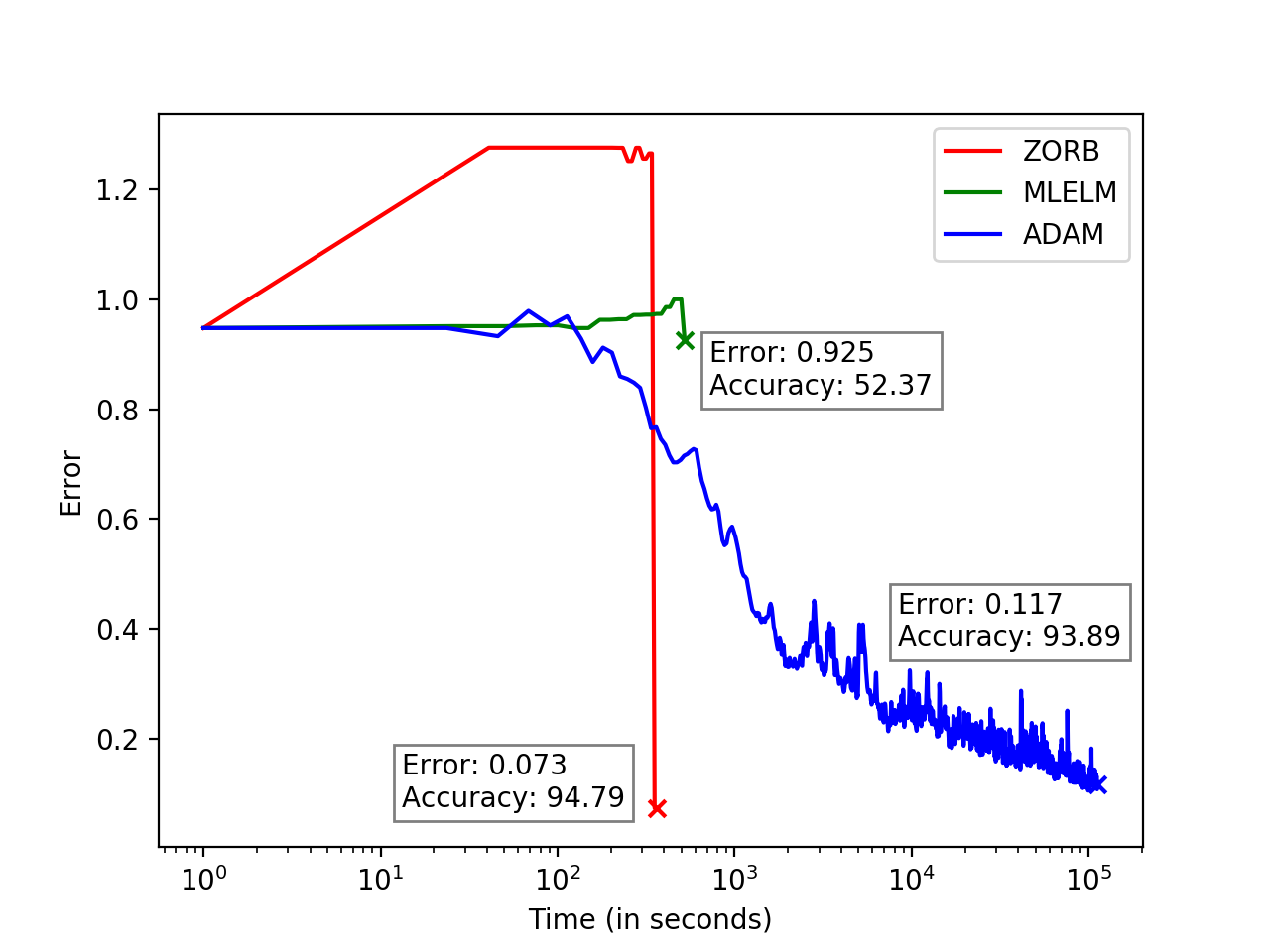}
}
\end{minipage}
\caption{Left: Three deep NNs with varying number of layers were trained on MNIST \citep{deng2012mnist}. ZORB trains NNs magnitudes faster than BP with Adam without degradation in performance.
Right: Training error vs Time (log-scaled) plot comparing three learning algorithms for the 11-layered NN.
}
\label{fig:mnist}
\end{figure}

\section{Zeroth-Order Relaxed Backpropagation (ZORB)}

In this section, we will present the Zeroth-Order Relaxed Backpropagation (ZORB) algorithm for NNs. 
We use the term ``Zeroth-Order'' because ZORB does not calculate the gradients of the error function. 
Additionally, ZORB is a ``Relaxed'' method in the sense that the inputs and outputs are fixed and we use iterative matrix computations to optimize the weights of the NN \citep{goffin1980relaxation, young2014iterative, varga62_iterat}.

\subsection{Training the network}
For didactic purposes, we start by addressing the problem of training a network that does not involve any activation functions.
The goal of ZORB is to match the predictions at each layer with targets provided by the layers above, implicitly reducing the sum of squared errors.
Let $\mathbf{X} \in \mathbb{R}^{d_{in} \times n}$ represent the input matrix, with $n$ being the number of samples and $d_{in}$ is the dimension of the input vector in a sample. 
Let $\mathbf{Y} \in \mathbb{R}^{ d_{out} \times n}$ represents the output matrix, where $d_{out}$ is the dimension of the output vector in a sample. 
The weights and biases of each hidden layer $l$ is represented by $\mathbf{W}_{l} \in \mathbb{R}^{d_{l} \times d_{l-1}}$ and $\mathbf{b}_{l} \in \mathbb{R}^{ d_{l} \times 1}$ respectively, where $d_{l-1}$ is the input dimension of a sample to layer $l$ and $d_{l}$ is the number of neurons in that layer. 
Then, a network $f$ with $L-1$ hidden layers and without activations can be written as $f(\mathbf{X}) = \mathbf{W}_L \mathbf{W}_{L-1} \dotso \mathbf{W}_1 \mathbf{X}$.

ZORB trains layers sequentially with the aim of generating an adequate representation of the input $X$ at layer $l-1$ before training layer $l$.
ZORB's forward pass is similar to the BP algorithm when no activation functions are used.
During the backward pass, a layer $l$ backpropagates information, as a feedback matrix $\mathbf{F}_{l-1}$, to layer $l-1$. 
This feedback is the expected matrix input that would allow the output layer to give correct predictions, without tuning it's parameters. 
Feedback $\mathbf{F}_{l-1}$ to layer $l-1$ can be calculated as 
\begin{equation}
    \mathbf{F}_{l-1} =  \mathbf{W}_{l}^{+} \times (\mathbf{F}_{l} - \mathbf{b}_{l})
\end{equation}
where $\mathbf{F}_{l}$ is the feedback matrix received by layer $l$, $\mathbf{b}_{l}$ is the bias vector of layer $l$, $\mathbf{W}_{l}$ is the weights of layer $l$ and $(\cdot)^{+}$ represents the Moore-Penrose pseudoinverse of a matrix.
If layer $l = L$ is the output layer, then $\mathbf{F}_{L} = \mathbf{Y}$.
$\mathbf{F}_{L-1}$ behaves as a target vector for the last hidden layer.
This process continues until a feedback matrix $F_1$ to the first hidden layer is calculated.
The parameters of first hidden layer are updated using the following equation:
\begin{equation}
    \begin{bmatrix}
    \mathbf{W}_{1} \\
    \mathbf{b}_{1}
    \end{bmatrix}
    =
   \mathbf{F}_{1} \times \mathbf{X}^{+}
\end{equation}
where $\mathbf{W}_{1}$ is the weight matrix for the first hidden layer, $\mathbf{b}_{1}$ is the bias vector for the first hidden layer, and $\mathbf{F}_{1}$ is the feedback received by the first hidden layer from the second hidden layer. 
Once the weights are updated, the input $\mathbf{X}$ is propagated through the first hidden layer. 
The resultant matrix now acts as the input to the rest of the network.
The whole process repeats until all layers are trained. 
The algorithmic form and analysis of ZORB is available in appendix \ref{sec:analysis}.

\subsection{Introducing non-linearity}
Activation functions allow for the non-linear projection of data.
ZORB does not require any derivatives and can use most activation functions.
The first forward propagation through an activation function is the same as the traditional BP algorithm. 
While backpropagating through activation functions, the inverse of the activation function is applied, i.e., the feedback matrices must be deactivated.
However, values in the feedback matrix may not lie in the range of the activation function, inhibiting invertibility.
Therefore, we introduce adaptive activation functions to store and apply linear scaling/shifting operations on the feedback matrix.
This allows the use activation functions while maintaining expressivity of NNs.
During the next forward propagation, activation functions reverse their linear corrections. 
Please see table \ref{tab:activations} for a list of common activation functions and how they are used in ZORB.

\begin{table}[t]
\centering
\resizebox{\columnwidth}{!}{
\begin{tabular}{cccccc}
 \textbf{Activation} & \textbf{Domain} & \textbf{Range} & \texttt{\textbf{activate(x)}} & \texttt{\textbf{deactivate(y)}} & \texttt{\textbf{activate(x)}} \\
 \textbf{Function} & & & \textbf{[without shift/scale]} & & \textbf{[with shift/scale]} \\ \\ \hline \\
 & & & & & \\
Linear & (-$\infty$, $\infty$) & (-$\infty$, $\infty$) & \texttt{return x} & \texttt{return y} & \texttt{return x} \\ 
 & & & & & \\ \\ \hline \\
 & & & &  \texttt{low = min(y)} & \texttt{return $\frac{1}{1 + e^{-x}}$} \\
Sigmoid & (-$\infty$, $\infty$) & (0, 1) & \texttt{return $\frac{1}{1 + e^{-x}}$} & \texttt{high = max(y)} &  \texttt{$\times$(high - low)} \\
 & & & & \texttt{return $log(\frac{y}{1 - y})$} & \texttt{+ low} \\ \\ \hline \\ 
 & & & &  \texttt{low = min(y)} & \texttt{return $\frac{e^{x} - e^{-x}}{e^{x} + e^{-x}}$}  \\
Tanh & (-$\infty$, $\infty$) & (-1, 1) & \texttt{return $\frac{e^{x} - e^{-x}}{e^{x} + e^{-x}}$} & \texttt{high = max(y)} & \texttt{$\times$(high - low)} \\
 & & & & \texttt{return $\frac{1}{2} \times log(\frac{1 + y}{1 - y})$} & \texttt{+ low} \\ \\ \hline \\ 
 & & & &  \texttt{if(y $<$ 0) then} &  \\
ReLU & (-$\infty$, $\infty$) & (0, $\infty$) & \texttt{return max(0, x)} & \texttt{return random(-1, 0)} & \texttt{return max(0, x)} \\
 & & & & \texttt{else return y} & \\ \\ \hline \\ 
 & & & \texttt{z = exp(x)} & \texttt{z = y $\times$ total} & \texttt{z = exp(x)} \\
Softmax & (-$\infty$, $\infty$)$\in R^{d}$ & (0, 1)$\in R^{d}$ & \texttt{total = sum(z)} & \texttt{total = 0} & \texttt{total = sum(z)} \\
 & & & \texttt{return $\frac{z}{total}$} & \texttt{return log(z)} & \texttt{return $\frac{z}{total}$} \\ \\ \hline
\end{tabular}
}
\caption{Activation functions and their linear correction operations.}
\label{tab:activations}
\end{table}

\subsection{Convolutional Neural Networks}

We now present ZORB in the context of Convolutional Neural Networks (CNNs) \citep{krizhevsky2012imagenet} by mimicking the functionality of a filter with a single neuron.
For a traditional convolutional layer, patches of the input are extracted that correspond to the filter being dragged across the input. 
These patches are multiplied with the filter weights element-wise and summed to a scalar value. 
For a ZORB convolutional layer, we reshape the patches and the filter weights so that they are vectors. 
Then, the scalar output can be obtained by the inner product between the reshaped patch and filter.
To invert this process we stack the reshaped vectors into a data matrix, and use the pseudoinverse to simultaneously approximate the inverse of the inner-product for each patch. 
When more than one filter is used, we average the feedback between the filters.
This leverages ZORB's original forward pass, backpropagation and update functions.
\begin{figure}[h]
    \centering
    \includegraphics[width=0.9\linewidth]{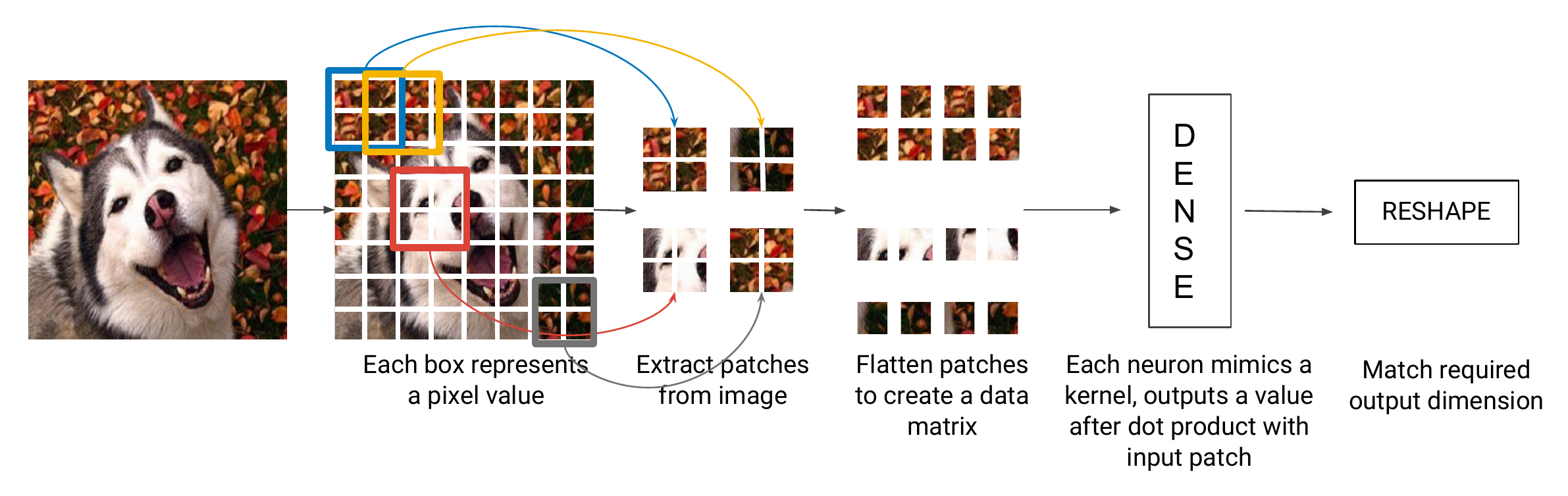}
    \caption{Representing Convolutional Neural Networks with ZORB. Patches of the image are extracted, reshaped and stacked in a matrix. ZORB is applied to learn the dense weight matrix which corresponds to the stacked reshaping of the filter weights.}
    \label{fig:zorbcnn}
\end{figure}

\section{Experiments and Results}

\begin{figure}[h]
\begin{minipage}[]{0.77\linewidth}
\centering
\resizebox{\columnwidth}{!}{
\begin{tabular}{|c|c|c|c|c|c|c|}
\hline
\textbf{Dataset} & \textbf{Training} & \multicolumn{2}{c|}{\textbf{Train}} & \multicolumn{2}{c|}{\textbf{Test}} & \textbf{Time} \\ \cline{3-6}
& \textbf{Algorithm} & \textbf{Error (MSE)} & \textbf{Accuracy (\%)} & \textbf{Error (MSE)} & \textbf{Accuracy (\%)} & \textbf{(s)} \\ \Xhline{3\arrayrulewidth}
Boston & Adam & $4.182\pm0.135$ & - & $4.405\pm\textbf{0.056}$ & - & $15.07\pm0.133$ \\ \cline{2-7}
Housing & MLELM & $3.109\pm0.102$ & - & $\textbf{3.543}\pm0.108$ & - & $0.028\pm0.003$ \\ \cline{2-7}
\citep{bollinger1981book} & ZORB & $\textbf{2.597}\pm\textbf{0.048}$ & - & $3.567\pm0.070$ & - & $\textbf{0.025}\pm\textbf{0.002}$ \\ \Xhline{3\arrayrulewidth}
& Adam & $0.043\pm0.003$ & - & $0.134\pm\textbf{0.021}$ & - & $122.4\pm2.862$ \\ \cline{2-7}
Sinc & MLELM & $\textbf{0.002}\pm\textbf{0.002}$ & - & $8\times10^4\pm2\times10^4$ & - & $\textbf{0.378}\pm0.038$ \\ \cline{2-7}
\citep{wang2011study} & ZORB & $0.015\pm0.006$ & - & $\textbf{0.128}\pm0.047$ & - & $0.414\pm\textbf{0.004}$ \\ \Xhline{3\arrayrulewidth}
& Adam & $0.075\pm\textbf{0.005}$ & $\textbf{97.90}\pm\textbf{0.83}$ & $0.079\pm\textbf{0.013}$ & $97.56\pm\textbf{1.20}$ & $4.087\pm0.069$ \\ \cline{2-7}
Iris & MLELM & $0.586\pm0.012$ & $95.71\pm1.49$ & $0.586\pm0.014$ & $96.00\pm3.27$ & $\textbf{0.004}\pm\textbf{0.001}$ \\ \cline{2-7}
\citep{fisher1936use} & ZORB & $\textbf{0.048}\pm0.010$ & $96.95\pm1.26$ & $\textbf{0.041}\pm0.021$ & $\textbf{98.22}\pm1.66$ & $0.005\pm\textbf{0.001}$ \\ \Xhline{3\arrayrulewidth}
& Adam & $\textbf{0.035}\pm0.015$ & $\textbf{97.73}\pm1.10$ & $0.052\pm0.015$ & $\textbf{97.48}\pm1.32$ & $17.425\pm3.339$ \\ \cline{2-7}
XOR & MLELM & $0.438\pm0.017$ & $59.97\pm3.00$ & $0.451\pm0.016$ & $54.75\pm2.89$ & $\textbf{0.048}\pm\textbf{0.003}$ \\ \cline{2-7}
\citep{ranganathan2018new} & ZORB & $0.039\pm\textbf{0.007}$ & $96.86\pm\textbf{0.81}$ & $\textbf{0.049}\pm\textbf{0.009}$ & $95.94\pm\textbf{1.02}$ & $0.068\pm0.006$ \\ \Xhline{3\arrayrulewidth}
Two & Adam & $\textbf{0.023}\pm0.064$ & $\textbf{96.18}\pm5.19$ & $\textbf{0.104}\pm0.067$ & $\textbf{92.58}\pm6.17$ & $7.459\pm3.789$ \\ \cline{2-7}
Spirals & MLELM & $0.495\pm\textbf{0.003}$ & $50.00\pm\textbf{0.00}$ & $0.496\pm\textbf{0.003}$ & $50.00\pm\textbf{0.00}$ & $\textbf{0.032}\pm\textbf{0.002}$ \\ \cline{2-7}
\citep{lang1988learning} & ZORB & $0.153\pm0.043$ & $84.89\pm3.91$ & $0.168\pm0.035$ & $83.50\pm3.94$ & $0.039\pm0.004$ \\ \Xhline{2\arrayrulewidth}
\end{tabular}
}
\end{minipage}
\begin{minipage}[]{0.22\linewidth}
\includegraphics[width=1\textwidth]{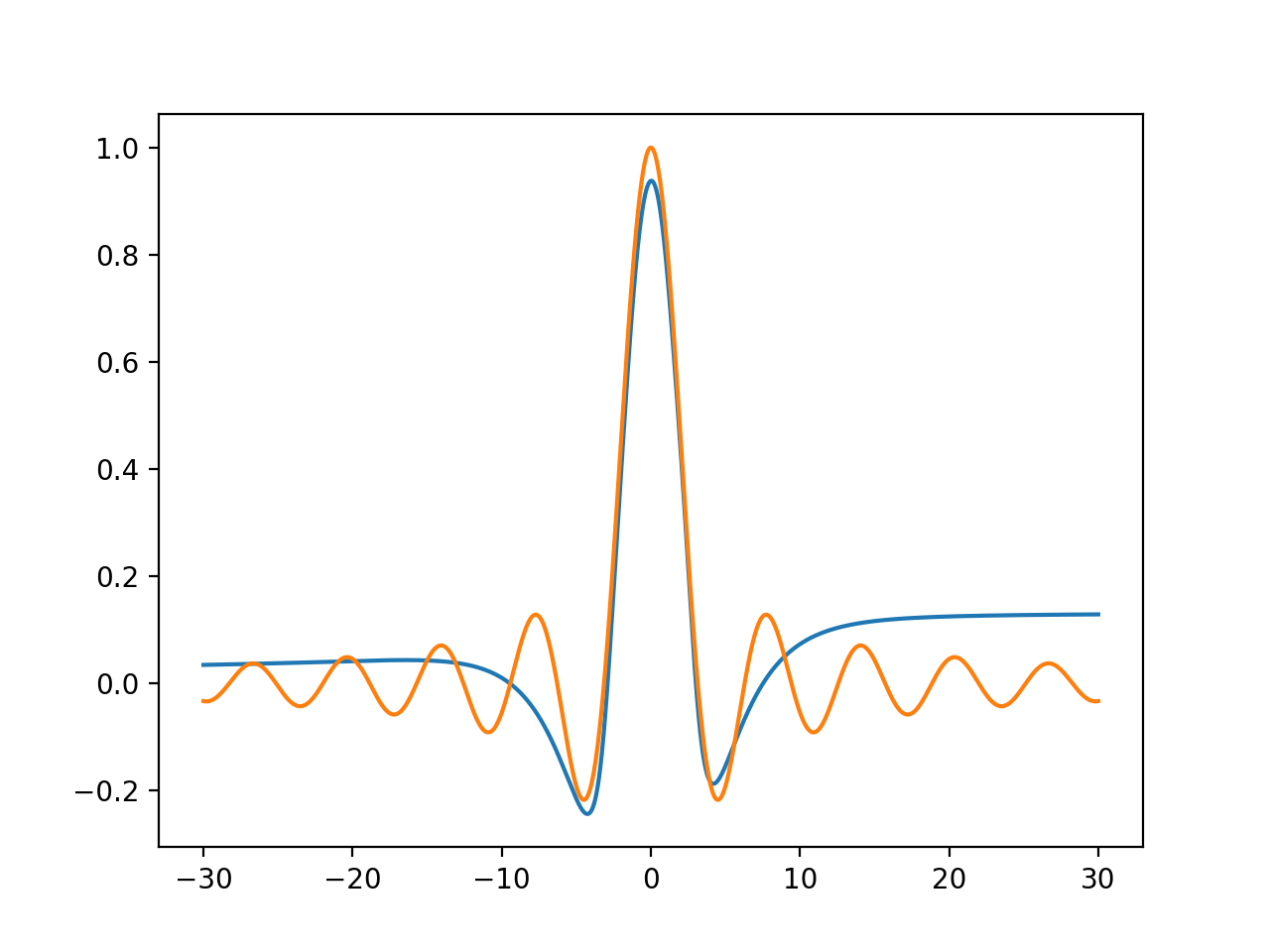}
\includegraphics[width=1\textwidth]{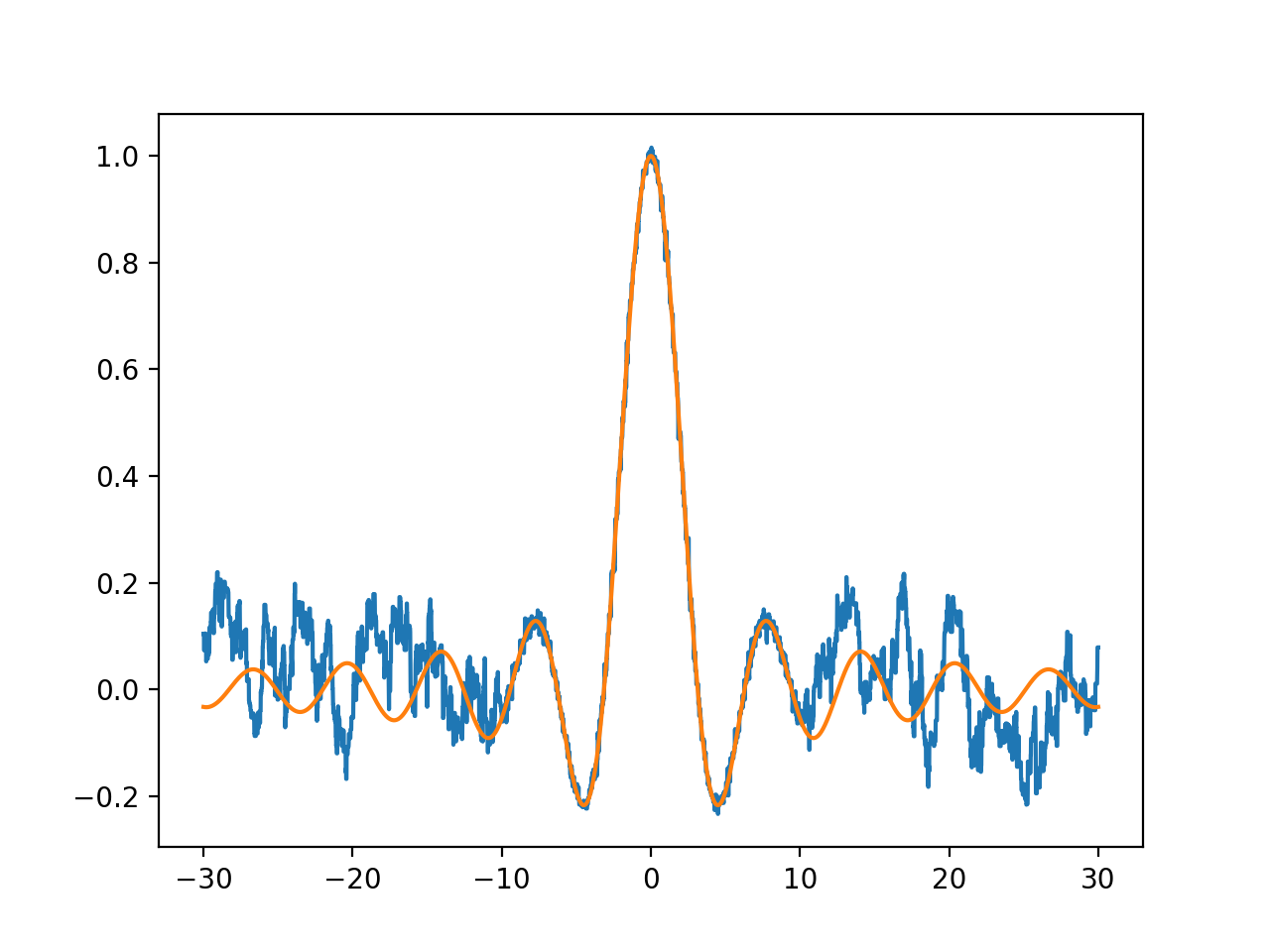}
\end{minipage}
\caption{Left: Performance of ZORB compared to MLELM and the BP with Adam. We observe that ZORB offers performance that is comparable to Adam, while training NNs magnitudes faster. Right: Visualization of predictions on the Sinc function. Orange: Ground truth values, Blue: Predicted values. Right Top: Adam, Right Bottom: ZORB.}
\label{fig:ffnn}
\end{figure}

We compare ZORB to BP with Adam \citep{kingma2014adam} and Multi-layered Extreme Learning Machines (ML-ELM) \citep{yang2015multilayer}.
For consistency in the optimization process across algorithms, we train feed-forward NNs to reduce the sum of squared error.
To compare training times, all algorithms were implemented using NumPy \citep{van2011numpy} and Autograd \citep{maclaurin2015autograd}.
Table from figure \ref{fig:ffnn} displays the results from benchmark experiments used to verify the working of NNs.
To test ZORB on a large dataset, we trained 3 NNs with varying number of layers on the MNIST dataset \citep{deng2012mnist}, displaying results in figure \ref{fig:mnist}.
ZORB is compared against baseline training algorithms on 3 metrics: Mean squared error, classification accuracy and wall-clock time.
The time displayed in tables for the Adam optimizer is minimum between the time taken to reach ZORB's error rate and time taken to reach a maximum iteration count.
Dataset statistics, network architectures and hyperparameter values have been provided in section \ref{sec:appendix_experiments} of the appendix.
Codes are publicly available at \url{https://github.com/varunranga/zorb-numpy}. 

Across all experiments, ZORB trains the NNs magnitudes faster than the BP algorithm with Adam without any compromise in training and testing performance.
ML-ELMs do not perform as well as ZORB or Adam, since the hidden layers are never trained. 
ZORB is able to match the training speed of ML-ELMs while achieving performance comparable to the BP algorithm.
We compare extrapolation abilities of the NNs trained to model the Sinc function \citep{wang2011study}.
The inputs in the training set ranged from $X \in [-10, +10]$, while the test set contained inputs ranging from $X \in [-30, -10) \cup (+10, +30]$. 
Figure \ref{fig:ffnn} (Right) visualizes this experiment. 
We observe that the Adam-trained network (Figure 2-Right Top) does not model the crests and troughs of the function beyond the range $[-10, +10]$. 
Instead, it simply maps values $[-30, -10) \cup (+10, +30]$ to constant values. 
In contrast, ZORB (Figure 2-Right Bottom) trained the network to recognize the non-linear oscillations outside the range $[-10, +10]$. 
We can see that ZORB-trained networks can generalize better than the benchmark algorithms, which is confirmed by the low error rate displayed the table of figure \ref{fig:ffnn}.

\begin{table}[h]
\centering
\resizebox{\columnwidth}{!}{
\begin{tabular}{|c?c|c|c|c?c|c|c|c?c|c|c|c|}
\hline
\textbf{\#S} & \multicolumn{4}{c?}{\textbf{Conv1}} & \multicolumn{4}{c?}{\textbf{Conv2}} & \multicolumn{4}{c|}{\textbf{Conv3}} \\ \cline{2-13} 
& \multicolumn{2}{c|}{\textbf{ZORB}} & \multicolumn{2}{c?}{\textbf{Adam}} & \multicolumn{2}{c|}{\textbf{ZORB}} & \multicolumn{2}{c?}{\textbf{Adam}} & \multicolumn{2}{c|}{\textbf{ZORB}} &
\multicolumn{2}{c|}{\textbf{Adam}} \\ \cline{2-13} 
&
\textbf{Acc.(\%)} & \multicolumn{1}{c|}{\textbf{Time(s)}} &
\textbf{Acc.(\%)} & \multicolumn{1}{c?}{\textbf{Time(s)}} &
\textbf{Acc.(\%)} & \multicolumn{1}{c|}{\textbf{Time(s)}} &
\textbf{Acc.(\%)} & \multicolumn{1}{c?}{\textbf{Time(s)}} &
\textbf{Acc.(\%)} & \multicolumn{1}{c|}{\textbf{Time(s)}} &
\textbf{Acc.(\%)} & \multicolumn{1}{c|}{\textbf{Time(s)}} \\ \Xhline{2.5\arrayrulewidth}
$10$ & $13.16$ & $\textbf{0.169}$ & $\textbf{15.14}$ & $136.2$ & $10.81$ & $\textbf{0.722}$ & $\textbf{15.24}$ & $1278$ & $11.07$ & $\textbf{2.006}$ & $\textbf{14.73}$ & $2520$ \\ \hline
$20$ & $15.78$ & $\textbf{0.283}$ & $\textbf{16.47}$ & $207.5$ & $11.99$ & $\textbf{1.361}$ & $\textbf{17.02}$ & $1183$ & $10.68$ & $\textbf{3.891}$ & $\textbf{16.26}$ & $3622$ \\ \hline
$40$ & $17.14$ & $\textbf{0.512}$ & $\textbf{18.45}$ & $520.5$ & $14.38$ & $\textbf{2.604}$ & $\textbf{17.07}$ & $1386$ & $13.04$ & $\textbf{7.420}$ & $\textbf{17.28}$ & $4860$ \\ \hline
$80$ & $20.25$ & $\textbf{1.056}$ & $\textbf{22.45}$ & $613.5$ & $17.98$ & $\textbf{5.197}$ & $\textbf{18.96}$ & $1874$ & $15.79$ & $\textbf{14.54}$ & $\textbf{18.15}$ & $6966$ \\ \hline
$160$ & $23.33$ & $\textbf{2.462}$ & $\textbf{24.32}$ & $592.5$ & $\textbf{22.89}$ & $\textbf{10.87}$ & $19.74$ & $1787$ & $\textbf{20.76}$ & $\textbf{30.09}$ & $19.70$ & $5503$ \\ \hline
$320$ & $\textbf{28.71}$ & $\textbf{4.688}$ & $26.44$ & $622.0$ & $\textbf{25.95}$ & $\textbf{22.23}$ & $22.48$ & $2296$ & $\textbf{25.68}$ & $\textbf{61.38}$ & $20.70$ & $3730$ \\ \hline
$640$ & $\textbf{32.37}$ & $\textbf{9.972}$ & $28.61$ & $748.0$ & $\textbf{28.55}$ & $\textbf{45.26}$ & $26.63$ & $1783$ & $\textbf{29.06}$ & $\textbf{125.6}$ & $22.98$ & $2192$ \\ \hline
$1280$ & $\textbf{33.30}$ & $\textbf{21.67}$ & $32.16$ & $836.5$ & $29.70$ & $\textbf{95.05}$ & $\textbf{33.12}$ & $2417$ & $\textbf{30.69}$ & $\textbf{262.6}$ & $28.43$ & $4324$ \\ \hline
\end{tabular}
}
\caption{Performance of CNNs trained by ZORB and BP with Adam on subsamples of CIFAR-10.}
\label{tab:cnn}
\end{table}

We also compare the performance of CNNs trained by ZORB against a standard 
implementation provided by Tensorflow \citep{abadi15_tensor} on a subsample of the CIFAR-10
dataset \citep{krizhevsky2009learning}. Similar to previous work exploring Neural
Tangent Kernels \citep{arora19_harnes_power_infin_wide_deep}, we train networks on
$n = \{10 ,\dotso, 1280\}$ samples from CIFAR-10 and evaluate over the entire
test set.
We train 3 CNNs with number of convolution layers $\in \{1,2,3\}$ with number of filters $\in \{32,64,128\}$.
Training CNNs using the BP algorithm involved reducing the crossentropy loss, since mean squared error minimization resulted in poor performance of CNNs \citep{rosasco2004loss}.
No change was made to ZORB's objective.
Referring to table \ref{tab:cnn}, we observe that ZORB continues to remain competitive to the BP algorithm in terms of accuracy while significantly improving training speed.
\vspace{-3mm}
\section{Related Work}
\vspace{-3mm}
\textbf{Extreme Learning Machines (ELMs)} \citep{huang2015trends, tapson2013learning, rossen1991closed} are relaxation based approaches \citep{goffin1980relaxation}, proposed to provide fast solutions to feed-forward networks.
ELMs and it's variants, including the Multi-layer ELMs (ML-ELMs) \citep{yang2015multilayer}, randomly project the input matrix to a higher dimensional space to fit the projected input to the required output using the pseudoinverse operation.
Although ELMs are a relatively new line of study, they are not competitive to NNs trained using ZORB or the BP algorithm. 
Randomly extracting input features reduces the training time, but does not provide the output layer with vital information that is ignored during the
randomization of weights in the hidden layer.

\textbf{Target Propagation} \citep{lee2015difference}
proposes a similar research direction to our work: find appropriate targets for each layer such that the global loss will reduce. 
In Target Propagation, a decoder network learns to perform an approximate inverse mapping. 
In ZORB, the true inverse of an adaptive activation function is supported by their linear corrective operations. 
The performance achieved by target propagation is similar to BP, but is much slower due to training a separate inverse network and with its own additional hyperparameters.

\textbf{Alternating Descent Method of Multipliers (ADMM)} \citep{taylor16_train_neural_networ_without_gradien} uses the least-squares objective to enable gradient-free optimization of NNs.
Both ZORB and ADMM algorithms employ a layer-wise training approach to solve a least-squares problem.
ADMM is formulated as a constrained optimization problem, whereas ZORB does not enforce constraints directly in the optimization problem.
In contrast to ZORB, ADMM requires numerous hyperparameters in the form of Lagrange multipliers.
The authors of ADMM focus on scalability to massive datasets via distributed computation. 
We aim to provide a solution towards reducing the development time of NNs.
\vspace{-3mm}
\section{Future Work and Conclusion}
\vspace{-3mm}
For future work, ZORB can be applied to non-differentiable operations, like novel activation functions and attention mechanisms \citep{vaswani2017attention}.
Next, a regularized pseudoinverse procedure proposed by Barata et al. \citep{barata2012moore} may be adopted to provide a self-regularized solution for the weights of a network.
ZORB could also provide a new direction for warm starting techniques \citep{ash2019difficulty}.
Griville's Theorem \citep{rao1972generalized} and the Bordering Algorithm \citep{claerbout1985fundamentals} may be used to efficiently recalculate the SVD and therefore, the pseudoinverse of matrices.
In particular, this would enable ZORB in the online learning regime.

This paper introduced a novel derivative-free training algorithm for neural networks called Zeroth-Order Relaxed Backpropagation (ZORB).
ZORB combines the rapid training of ELMs with the accuracy of BP.
Experiments have been conducted on several datasets using both feed-forward and convolution networks. 
Results from these experiments verify that ZORB is competitive to the BP algorithm in terms of accuracy while significantly reducing training time.


\small

\section*{Acknowledgements}
We would like to thank Natarajan Subramanyam and Martha White for preliminary discussions and directions.

\newpage
\bibliographystyle{unsrt}
\bibliography{example_paper}

\begin{thebibliography}{}

\bibitem[Abadi et~al., 2015]{abadi15_tensor}
Abadi, M., Agarwal, A., Barham, P., Brevdo, E., Chen, Z., Citro, C., Corrado,
  G.~S., Davis, A., Dean, J., Devin, M., Ghemawat, S., Goodfellow, I., Harp,
  A., Irving, G., Isard, M., Jia, Y., Jozefowicz, R., Kaiser, L., Kudlur, M.,
  Levenberg, J., Man\'{e}, D., Monga, R., Moore, S., Murray, D., Olah, C.,
  Schuster, M., Shlens, J., Steiner, B., Sutskever, I., Talwar, K., Tucker, P.,
  Vanhoucke, V., Vasudevan, V., Vi\'{e}gas, F., Vinyals, O., Warden, P.,
  Wattenberg, M., Wicke, M., Yu, Y., and Zheng, X. (2015).
\newblock {TensorFlow}: Large-scale machine learning on heterogeneous systems.
\newblock Software available from tensorflow.org.

\bibitem[Arora et~al., 2019]{arora19_harnes_power_infin_wide_deep}
Arora, S., Du, S.~S., Li, Z., Salakhutdinov, R., Wang, R., and Yu, D. (2019).
\newblock Harnessing the power of infinitely wide deep nets on small-data
  tasks.
\newblock {\em CoRR}.

\bibitem[Ash and Adams, 2019]{ash2019difficulty}
Ash, J.~T. and Adams, R.~P. (2019).
\newblock On the difficulty of warm-starting neural network training.
\newblock {\em arXiv preprint arXiv:1910.08475}.

\bibitem[Barata and Hussein, 2012]{barata2012moore}
Barata, J. C.~A. and Hussein, M.~S. (2012).
\newblock The moore--penrose pseudoinverse: A tutorial review of the theory.
\newblock {\em Brazilian Journal of Physics}, 42(1-2):146--165.

\bibitem[Bergstra and Bengio, 2012]{bergstra2012random}
Bergstra, J. and Bengio, Y. (2012).
\newblock Random search for hyper-parameter optimization.
\newblock {\em Journal of Machine Learning Research}, 13(Feb):281--305.

\bibitem[Bollinger, 1981]{bollinger1981book}
Bollinger, G. (1981).
\newblock Book review: Regression diagnostics: Identifying influential data and
  sources of collinearity.

\bibitem[Cai, 2019]{cai2019tutorial}
Cai, X. L. S. W.~Y. (2019).
\newblock Tutorial: Complexity analysis of singular value decomposition and its
  variants.
\newblock {\em arXiv preprint arXiv:1906.12085}.

\bibitem[Claerbout, 1985]{claerbout1985fundamentals}
Claerbout, J.~F. (1985).
\newblock {\em Fundamentals of geophysical data processing}.
\newblock Citeseer.

\bibitem[Deng, 2012]{deng2012mnist}
Deng, L. (2012).
\newblock The mnist database of handwritten digit images for machine learning
  research [best of the web].
\newblock {\em IEEE Signal Processing Magazine}, 29(6):141--142.

\bibitem[Fisher, 1936]{fisher1936use}
Fisher, R.~A. (1936).
\newblock The use of multiple measurements in taxonomic problems.
\newblock {\em Annals of eugenics}, 7(2):179--188.

\bibitem[Gaivoronski, 1994]{gaivoronski1994convergence}
Gaivoronski, A.~A. (1994).
\newblock Convergence properties of backpropagation for neural nets via theory
  of stochastic gradient methods. part 1.
\newblock {\em Optimization methods and Software}, 4(2):117--134.

\bibitem[Goffin, 1980]{goffin1980relaxation}
Goffin, J.-L. (1980).
\newblock The relaxation method for solving systems of linear inequalities.
\newblock {\em Mathematics of Operations Research}, 5(3):388--414.

\bibitem[Golan, 2012]{golan2012moore}
Golan, J.~S. (2012).
\newblock Moore--penrose pseudoinverses.
\newblock pages 441--452.

\bibitem[Guo et~al., 2001]{guo2001pseudoinverse}
Guo, P., Lyu, M.~R., and Mastorakis, N. (2001).
\newblock Pseudoinverse learning algorithm for feedforward neural networks.
\newblock {\em Advances in Neural Networks and Applications}, 1(321-326).

\bibitem[Hinton et~al., 2012]{hinton2012neural}
Hinton, G., Srivastava, N., and Swersky, K. (2012).
\newblock Neural networks for machine learning lecture 6a overview of
  mini-batch gradient descent.
\newblock {\em Cited on}, 14:8.

\bibitem[Huang et~al., 2015]{huang2015trends}
Huang, G., Huang, G.-B., Song, S., and You, K. (2015).
\newblock Trends in extreme learning machines: A review.
\newblock {\em Neural Networks}, 61:32--48.

\bibitem[Kingma and Ba, 2014]{kingma2014adam}
Kingma, D.~P. and Ba, J. (2014).
\newblock Adam: A method for stochastic optimization.
\newblock {\em arXiv preprint arXiv:1412.6980}.

\bibitem[Krizhevsky et~al., 2009]{krizhevsky2009learning}
Krizhevsky, A., Hinton, G., et~al. (2009).
\newblock Learning multiple layers of features from tiny images.
\newblock Technical report, Citeseer.

\bibitem[Krizhevsky et~al., 2012]{krizhevsky2012imagenet}
Krizhevsky, A., Sutskever, I., and Hinton, G.~E. (2012).
\newblock Imagenet classification with deep convolutional neural networks.
\newblock In {\em Advances in neural information processing systems}, pages
  1097--1105.

\bibitem[Lang and Witbrock, 1988]{lang1988learning}
Lang, K.~J. and Witbrock, M.~J. (1988).
\newblock Learning to tell two spirals apart.
\newblock In {\em Proceedings of the 1988 connectionist models summer school},
  number 1989 in 1, pages 52--59. San Mateo.

\bibitem[Lee et~al., 2015]{lee2015difference}
Lee, D.-H., Zhang, S., Fischer, A., and Bengio, Y. (2015).
\newblock Difference target propagation.
\newblock In {\em Joint european conference on machine learning and knowledge
  discovery in databases}, pages 498--515. Springer.

\bibitem[Maclaurin et~al., 2015]{maclaurin2015autograd}
Maclaurin, D., Duvenaud, D., and Adams, R.~P. (2015).
\newblock Autograd: Effortless gradients in numpy.
\newblock In {\em ICML 2015 AutoML Workshop}, volume 238.

\bibitem[Nesterov, 1983]{nesterov1983method}
Nesterov, Y.~E. (1983).
\newblock A method for solving the convex programming problem with convergence
  rate o (1/k\^{} 2).
\newblock In {\em Dokl. akad. nauk Sssr}, volume 269, pages 543--547.

\bibitem[Nishihara et~al., 2017]{nishihara2017real}
Nishihara, R., Moritz, P., Wang, S., Tumanov, A., Paul, W., Schleier-Smith, J.,
  Liaw, R., Niknami, M., Jordan, M.~I., and Stoica, I. (2017).
\newblock Real-time machine learning: The missing pieces.
\newblock In {\em Proceedings of the 16th Workshop on Hot Topics in Operating
  Systems}, pages 106--110.

\bibitem[Penrose, 1956]{penrose1956best}
Penrose, R. (1956).
\newblock On best approximate solutions of linear matrix equations.
\newblock In {\em Mathematical Proceedings of the Cambridge Philosophical
  Society}, volume~52, pages 17--19. Cambridge University Press.

\bibitem[Qian, 1999]{qian1999momentum}
Qian, N. (1999).
\newblock On the momentum term in gradient descent learning algorithms.
\newblock {\em Neural networks}, 12(1):145--151.

\bibitem[Radivojac and White, 2019]{radivojac2019machine}
Radivojac, P. and White, M. (2019).
\newblock Machine learning handbook.

\bibitem[Ranganathan and Natarajan, 2018]{ranganathan2018new}
Ranganathan, V. and Natarajan, S. (2018).
\newblock A new backpropagation algorithm without gradient descent.
\newblock {\em arXiv preprint arXiv:1802.00027}.

\bibitem[Rao et~al., 1972]{rao1972generalized}
Rao, C.~R., Mitra, S.~K., et~al. (1972).
\newblock Generalized inverse of a matrix and its applications.
\newblock In {\em Proceedings of the Sixth Berkeley Symposium on Mathematical
  Statistics and Probability, Volume 1: Theory of Statistics}. The Regents of
  the University of California.

\bibitem[Rosasco et~al., 2004]{rosasco2004loss}
Rosasco, L., Vito, E.~D., Caponnetto, A., Piana, M., and Verri, A. (2004).
\newblock Are loss functions all the same?
\newblock {\em Neural Computation}, 16(5):1063--1076.

\bibitem[Rossen, 1991]{rossen1991closed}
Rossen, M.~L. (1991).
\newblock Closed-form inversion of backpropagation networks: theory and
  optimization issues.
\newblock In {\em Advances in Neural Information Processing Systems}, pages
  868--872.

\bibitem[Ruder, 2016]{ruder2016overview}
Ruder, S. (2016).
\newblock An overview of gradient descent optimization algorithms.
\newblock {\em arXiv preprint arXiv:1609.04747}.

\bibitem[Rumelhart et~al., 1986]{rumelhart1986learning}
Rumelhart, D.~E., Hinton, G.~E., and Williams, R.~J. (1986).
\newblock Learning representations by back-propagating errors.
\newblock {\em nature}, 323(6088):533--536.

\bibitem[Tapson and van Schaik, 2013]{tapson2013learning}
Tapson, J. and van Schaik, A. (2013).
\newblock Learning the pseudoinverse solution to network weights.
\newblock {\em Neural Networks}, 45:94--100.

\bibitem[Taylor et~al., 2016]{taylor16_train_neural_networ_without_gradien}
Taylor, G., Burmeister, R., Xu, Z., Singh, B., Patel, A., and Goldstein, T.
  (2016).
\newblock Training neural networks without gradients: A scalable admm approach.
\newblock {\em CoRR}.

\bibitem[Van Der~Walt et~al., 2011]{van2011numpy}
Van Der~Walt, S., Colbert, S.~C., and Varoquaux, G. (2011).
\newblock The numpy array: a structure for efficient numerical computation.
\newblock {\em Computing in Science \& Engineering}, 13(2):22.

\bibitem[Varga, 1962]{varga62_iterat}
Varga, R.~S. (1962).
\newblock {\em Iterative analysis}.
\newblock Springer.

\bibitem[Vaswani et~al., 2017]{vaswani2017attention}
Vaswani, A., Shazeer, N., Parmar, N., Uszkoreit, J., Jones, L., Gomez, A.~N.,
  Kaiser, {\L}., and Polosukhin, I. (2017).
\newblock Attention is all you need.
\newblock In {\em Advances in neural information processing systems}, pages
  5998--6008.

\bibitem[Wang et~al., 2011]{wang2011study}
Wang, Y., Cao, F., and Yuan, Y. (2011).
\newblock A study on effectiveness of extreme learning machine.
\newblock {\em Neurocomputing}, 74(16):2483--2490.

\bibitem[Yang and Wu, 2015]{yang2015multilayer}
Yang, Y. and Wu, Q.~J. (2015).
\newblock Multilayer extreme learning machine with subnetwork nodes for
  representation learning.
\newblock {\em IEEE transactions on cybernetics}, 46(11):2570--2583.

\bibitem[Young, 2014]{young2014iterative}
Young, D.~M. (2014).
\newblock {\em Iterative solution of large linear systems}.
\newblock Elsevier.

\end{thebibliography}

\newpage
\appendix
\section*{Appendix}
\renewcommand{\thesection}{\Alph{section}}

\section{Preliminaries}

\subsection{Gradient Descent}
Gradient Descent \cite{ruder2016overview} is an iterative optimization algorithm to find a local minima of an error function surface.
It is based on the intuition that if the multivariable cost function $c(w_0)$ is defined and differentiable in a neighborhood of a point $w_0$, then $c(w)$ decreases fastest if one goes from $w_0$ in the direction of the negative gradient of $c$ at $w_0$, that is, $- \nabla c(w_0)$.
It involves approximating the function using the Taylor series, given by:
\begin{equation}
c(w) = \Sigma_{n=0}^{\infty} \frac{c^{(n)}(w_0)}{n!}(w - w_0)^n
\end{equation}
where $c^{(n)}(w_0)$ is the $n^{th}$ derivative of the function $c(w)$ evaluated at point $w_0$.
Newton's method uses a second-order approximation of this Taylor series by using only the first three terms.
A stationary point of the approximated error function can be found by setting the derivative of the function to zero and solving for the required parameters.
\begin{equation}
c'(w) \approx c'(w_0) + (w - w_0) c''(w_0) = 0
\end{equation}
Upon solving for the parameter $w$, we arrive at the following expression.
\begin{equation}
w = w_0 - \frac{c'(w_0)}{c''(w_0)}
\end{equation}
Since this parameter update is based off an approximation of the error function, it is iteratively updated based on the current parameter values.
This requires the recomputation of both the first derivative and the second derivative of the function.
Compared to computing the first derivative, computing the second derivative is quite expensive.
Instead, one approximates the second-order derivative using a constant value called step size or le   arning rate \cite{radivojac2019machine}.
The first-order GD update rule is given by substituting a learning rate in place of second derivative information $\mu \approx c''(w_{0})$.

There are several variants of the GD that has been applied in the context of neural networks.
Batch Gradient Descent (BGD) involves updating the weights of the network after accumulating the gradients calculated for every sample in the dataset.
Although this procedure is guaranteed to reach the local minima of a smooth convex error function, it has slower convergence rate since the gradients have to be computed for each sample in the dataset before updating the network.
Stochastic Gradient Descent (SGD) uses a single sample to approximate the gradient.
This results in faster convergence, but is certainly not the best approach to solving the problem.
Since SGD approximates the gradient using a single example, a noisier gradient is calculated, which can bump the weights out of a local minima.
Mini-batch Gradient Descent \cite{hinton2012neural} combines both the ideas by using a subset of the dataset to calculate the gradient.

\subsection{Moore-Penrose Pseudoinverse}
The pseudoinverse $A^{+}$ of a matrix $A$ is a generalization of the inverse matrix.
The most widely known type of matrix pseudoinverse is the Moore-Penrose Inverse.
A common use of the pseudoinverse operation is to find the minimum (Euclidean) norm solution to a system of linear equations with multiple solutions \cite{barata2012moore}.
To leverage parallelization, it is usually computed using the singular value decomposition.
The pseudoinverse operation allows us to find a set of values for variables involved in a equation. 
These values correspond to a local minima while reducing the mean squared error loss. 

\section{Analysis of ZORB}
\label{sec:analysis}

\begin{algorithm}[]
  \caption{ZORB Algorithm for Neural Networks}
  \label{alg:zorb}
  \begin{algorithmic}
    \STATE {\bfseries Input:} Input $\mathbf{X} \in \mathbb{R}^{d_{in} \times n}$,
    Output $\mathbf{Y} \in \mathbb{R}^{d_{out} \times n}$, Neural Network $\mathbf{N}$ with $L$ layers
    \FOR{$i = 1$ to $L$} \IF{$\mathbf{N}.layers[i]$ is a neural layer}
    \STATE $\mathbf{X} \leftarrow$ Concatenate $\mathbbm{1} \in R^{1 \times n}$ to $\mathbf{X}$ \COMMENT{To model bias}
    \STATE Forward propagate $\mathbf{X}$ through the network \COMMENT{To set scaling}
    \STATE $\mathbf{F} \leftarrow \mathbf{Y}$ \FOR{$j = L$ to $i+1$}
    \IF{$\mathbf{N}.layers[j]$ is a neural layer} \STATE
    $\mathbf{F} \leftarrow \mathbf{N}.\text{layers}[j].\mathbf{W}^+ \times (\mathbf{F} - \mathbf{N}.layers[j].\mathbf{b}) $
    \ELSIF{$\mathbf{N}.layers[j]$ is an activation layer} \STATE
    $\mathbf{F} \leftarrow \mathbf{N}.layers[j].deactivate(\mathbf{F})$ \ENDIF \ENDFOR \STATE
    $\begin{bmatrix} \mathbf{N}.layers[i].\mathbf{W} \\ \mathbf{N}.layers[i].\mathbf{b} \end{bmatrix} \leftarrow \mathbf{F} \times \mathbf{X}^{+} $ \COMMENT{Weight update}
    \STATE
    $\mathbf{X} \leftarrow\begin{bmatrix} \mathbf{N}.layers[i].\mathbf{W} \\ \mathbf{N}.layers[i].\mathbf{b} \end{bmatrix} \times  \mathbf{X} $ \COMMENT{Forward propagation}
    \ELSIF{$\mathbf{N}.layers[i]$ is an activation layer} \STATE
    $\mathbf{X} \leftarrow \mathbf{N}.layers[i].activate(\mathbf{X})$ \ENDIF \ENDFOR
  \end{algorithmic}
\end{algorithm}

In this section, we analyze ZORB's learning procedure. We begin by highlighing
some properties of the Moore-Penrse pseudoinverse. One fundamental property of
the Moore-Penrose pseudoinverse is its least squares minimizing property. Let
$\mathbf{W} \in \mathbb{R}^{k\times d}$,
$\mathbf{x} \in \mathbb{R}^{d \times 1}$ and
$\mathbf{y} \in \mathbb{R}^{k \times 1}$, then it is well known that
$\|\mathbf{Wx} - \mathbf{y} \|_{2} > \|\mathbf{Wz} - \mathbf{y} \|_{2}$ for
$\mathbf{z} = \mathbf{yW^{+}}$ where $\mathbf{W^{+}}$ is the pseudoinverse of
$\mathbf{W}$ \citep{penrose1956best, barata2012moore}. This result can be
extended to a batch of inputs by representing the input as a matrix
$\mathbf{X} \in \mathbb{R}^{d \times n}$ and
$\mathbf{Y} \in \mathbb{R}^{k \times n}$. Now, for the Frobenius norm
$\|\mathbf{A}\|_{F} = \sqrt{\sum_{i,j} [\mathbf{A}]_{ij}}$ we have that,
$\|\mathbf{WX} - \mathbf{Y} \|_{F} > \|\mathbf{WZ} - \mathbf{Y} \|_{F}$ for
$\mathbf{Z} = \mathbf{YW^{+}}$. If we consider the supervised learning setting where $\mathbf{X}$
fixed, and are estimating the weight matrix $\mathbf{W}$ then a similar result
also holds.
$\|\mathbf{WX} - \mathbf{Y} \|_{F} > \|\mathbf{ZX} - \mathbf{Y} \|_{F}$ where,
in the equation we now have that $\mathbf{Z} = \mathbf{YX^{+}}$.

\subsection{Optimization objective}
We first generalize the Moore-Penrose results above to ZORB applied to a neural network with zero hidden layers (perceptron) and no biases.
In this case, the output of the perceptron is characterized by $f(\mathbf{X} ; \mathbf{W}) = h(\mathbf{WX})$ where $\mathbf{W} \in \mathbb{R}^{k \times d}$ is the weight matrix, $\mathbf{X} \in \mathbb{R}^{d \times n}$ is a matrix corresponding to $n$ samples of a $d$-dimensional input and $h(\cdot)$ is an element-wise non-linearity.
Then, we can de-activate the $k$-dimensional target vector $\mathbf{Y} \in \mathbb{R}^{k \times n}$ to obtain the linear system $\mathbf{WX} \approx h^{-1}(\mathbf{Y})$.
The Frobenius norm is minimized by $\hat{\mathbf{W}} = h^{-1}(\mathbf{Y})\mathbf{X}^{+}$, which is the solution found by ZORB.

To formulate ZORB as an optimization problem, we write it as a layer-wise
objective with respect to the Frobenius norm.
We now denote the feedback matrix of an $L$-layer neural network $f(\mathbf{X}; \mathbf{W}_1, \dotso, \mathbf{W}_L)$ as $f^{-l}(\mathbf{Y}) = \mathbf{W}_{l+1}^+f^{-(l + 1)}(\mathbf{Y})$, with initial condition $f^{-L}(\mathbf{Y}) = \mathbf{Y}$. We also denote the activations from the previous hidden layer $f^{l}(\mathbf{X}) = h\big(\mathbf{W}_{l} f^{l-1}(\mathbf{X})\big)$, with initial condition $f^{0}(\mathbf{X}) = \mathbf{X}$.
Then, ZORB can be interpreted as the following layer-wise optimization problem:
\begin{equation}
\mathbf{W}^{l} = \argmin_{\mathbf{W}_{l}} \| \mathbf{W}_{l} f^{l-1}(\mathbf{X}) - f^{-l} (\mathbf{Y}) \|_{F}
\end{equation}

For feed-forward neural networks with one hidden layer, the output is characterized by $f(\mathbf{X} ; \mathbf{W}_1, \mathbf{W}_2) = h(\mathbf{W}_{2}h(\mathbf{W_{1}X}))$ for weight matrices $\mathbf{W}_{1}, \mathbf{W}_{2}$.
Then the optimized weight matrices are $\mathbf{\hat{W}_{1}} = h^{-1}\big(\mathbf{W}_{2}^{+} h^{-1}(\mathbf{Y})\big )\mathbf{X}^{+}$ and $\mathbf{\hat{W}_2} = h^{-1}(\mathbf{Y}) \left(h(\mathbf{\hat{W}_{1}X})\right)^{+}$, with the domain and range normalized as in Table \ref{tab:activations}.
The result for two or more hidden layers follows from induction.

\subsection{Complexity analysis}

In this subsection, we will discuss the time and space complexities of ZORB.
Let $N$ be the number of training samples, $L$ be the number of layers in the network and $M$ be the maximum of all input dimensions to every layer.
We assume that activation functions do not contribute to the time and space complexity.

\textbf{Time Complexity}: ZORB trains each layer iteratively.
There are $L$ iterations, which involves one forward pass, one backward pass and one update operation.
During the forward pass, the input $X$ goes through $L$ layers, invoking $L$ matrix multiplications.
The complexity of each matrix multiplication is $\mathcal{O}(NM^{2})$, hence the total complexity of one forward pass is $\mathcal{O}(LNM^{2})$.
During the backward pass, the feedback is backpropagated through $L-1$ layers and one layer is updated.
The operation $W_{L}^{+}$, which involves the calculation of the pseudoinverse of the matrix, is dominated by the SVD operation.
This operation has a time complexity of $\mathcal{O}(M^{3})$ \cite{cai2019tutorial}.
The calculation of the feedback matrix involves another matrix multiplication with time complexity $\mathcal{O}(NM^{2})$.
Therefore, the backpropagation process has a time complexity $\mathcal{O}(L(NM^{2} + M^{3}))$.
The update operation involves the pseudoinverse of the input matrix and a matrix multiplication, each having a time complexity of $\mathcal{O}(NM^{2})$.
The total time complexity for ZORB is $\mathcal{O}(L^{2}M^{2}(N + M))$.

\textbf{Space Complexity}: In our space complexity analysis, we include the space taken by the parameters of the network and the data.
The parameters have a space complexity of $\mathcal{O}(LM^{2})$, while
the data and its subsequent representations through the network take up $\mathcal{O}(NM)$ space.
The pseudoinverse operation used to calculate the feedback matrix at each backward step has a space complexity of $\mathcal{O}(M^{2})$ \cite{cai2019tutorial}.
Since a feedback matrix is used only by the particular layer, we can overwrite this space with a new feedback matrix.
The pseudoinverse operation used to update the layer utilizes $\mathcal{O}(NM)$ space.
The total space complexity of ZORB is $\mathcal{O}(M(LM + N))$.

\section{Experiment Details}
\label{sec:appendix_experiments}

In this section, information is provided for the reproducibility of the experiments conducted in the paper. The codes that are used to perform the experiments are publicly available at \url{https://github.com/varunranga/zorb-numpy}.

\subsection{Dataset Statistics}
Several benchmarking datasets were used to verify the working of ZORB. Two datasets for regression were used: Boston Housing \cite{bollinger1981book} (Linear Regression) and the Sinc function \cite{wang2011study} (Non-linear Regression). Three small scale datasets were used for classification: Iris \cite{fisher1936use} (Linear Classification), XOR \cite{ranganathan2018new} (Non-linear Classification) and the Two Spirals task \cite{lang1988learning} (Non-linear Classification). The MNIST dataset \cite{deng2012mnist} was used for large-scale classification. CIFAR-10 dataset \cite{krizhevsky2009learning} was used for training and evaluation of convolutional neural networks.
Table \ref{tab:dataset} presents the statistics related to the datasets.
\begin{table}[]
\centering
\begin{tabular}{|c|c|c|c|c|c|}
\hline
\textbf{Dataset} & \textbf{Type of} & \textbf{Input} & \textbf{Output} & \textbf{\# Training} & \textbf{\# Testing} \\
\textbf{} & \textbf{Task} & \textbf{Dim.} & \textbf{Dim.} & \textbf{samples} & \textbf{samples} \\ \hline
Boston Housing & Regression & 13 & 1 & 404 & 102 \\ \hline
Iris & Classification & 4 & 3 & 105 & 45 \\ \hline
Sinc & Regression & 1 & 1 & 2,001 & 6,001 \\ \hline
XOR & Classification & 2 & 1 & 1,000 & 1,000 \\ \hline
TwoSpirals & Classification & 2 & 1 & 280 & 120 \\ \hline
MNIST & Classification & 784 & 10 & 60,000 & 10,000 \\ \hline
CIFAR-10 & Classification & (32, 32, 3) & 10 & 50,000 & 10,000 \\ \hline
\end{tabular}
\caption{Statistics of datasets used in experiments.}
\label{tab:dataset}
\end{table}

\subsection{Dependencies}

\subsubsection{Hardware Dependencies}
The system used to perform experiments on feed-forward neural networks had the following specifications:
\begin{itemize}
    \item CPU : Intel(R) Core(TM) i7-6850K CPU @ 3.60GHz
    \item GPU (did not use) : 2 x Nvidia GeForce GTX 1080 Ti
    \item RAM : 64GB
    \item Hard disk space : 3.5TB
\end{itemize}

\subsubsection{Software Dependencies}
Table \ref{tab:sw} provides a list of software dependencies that are required to run the required codes. Please install these dependencies before running experiments.
\begin{table}[h]
\centering
\begin{tabular}{cc}
\hline
\textbf{Package / Software} & \textbf{Version} \\ \hline
Ubuntu & 18.04 \\
Python3 & 3.6.9 \\
Numpy & 1.18.1 \\
Tensorflow & 2.1.0 \\
Keras & 2.3.1 \\
Scikit-learn & 0.22.1 \\
Pandas & 0.25.3 \\
Autograd & 1.3 \\
Matplotlib & 3.1.1 \\
Pickle & (Available with Python3) \\
Argparse & (Available with Python3) \\
Time & (Available with Python3) \\
\hline
\end{tabular}
\caption{Software dependencies for running provided codes.}
\label{tab:sw}
\end{table}

\subsection{Running experiments}

\subsubsection{Feed-forward Neural Networks}

To reproduce the results as obtained in the paper, please run the bash script \texttt{run.sh} in the \texttt{Experiments/FFNN} folder. If you would like to create new initialization and perform the experiments, please run the bash script \texttt{init\_and\_run.sh} in the \texttt{Experiments/FFNN} folder. The script \texttt{show\_results.sh} in the folders \texttt{Experiments/FFNN/ZORB}, \texttt{Experiments/FFNN/MLELM} and \texttt{Experiments/FFNN/ADAM} extract results from the log files.

To train a network on a particular dataset using an algorithm, please run the python script \texttt{main.py} in the folder dedicated to the algorithm. Use the flag \texttt{-n} to specify the network architecture, and flag \texttt{-d} to specify the dataset. Use the \texttt{-h} flag for help.

The python scripts \texttt{sinc.py}, \texttt{xor.py} and \texttt{two\_spirals.py} visualize the network's performance on the three datasets. The python script \texttt{adam\_times.py} gets the time taken by the Adam optimizer to reach ZORB's train error or a maximum number of iterations, whichever is smaller.

\subsubsection{Convolutional Neural Networks}

To reproduce results as obtained in the paper for Convolutional Neural Networks, please run the bash scripts \texttt{run0.sh}, \texttt{run1.sh} and \texttt{run2.sh} in the folder \texttt{Experiments/CNN}. Instruction related to training a convolution neural network follow from the previous subsection.

\subsection{Architectures and Hyperparameters}

\subsubsection{Feed-forward Neural Networks}

Hyperparameter tuning is crucial for the performance of the Adam optimizer. Table \ref{tab:hyperparameters} provides the optimal values for the hyperparameters for a particular dataset and the network architecture as described in the paper. The number of samples per batch was kept constant at 32. For small-scale datasets, the number of iterations of the backpropagation algorithm was set to 2500, whereas for MNIST (a large-scale dataset), the number of iterations was set to 5000. Due to time constraints, the number of runs for the MNIST experiments was 1. The learning rate was tuned for values in $\{0.01, 0.005, 0.001, 0.0005, 0.0001\}$. The neural architecture used for each dataset is as follows:

\begin{enumerate}
    \item \textbf{Boston Housing}:
        \begin{itemize}
            \item Number of hidden layers: 1
            \item Number of neurons in each hidden layer: 32
            \item Activation after each hidden layer: sigmoid
            \item Number of neurons in the output layer: 1
            \item Activation after the output layer: linear
        \end{itemize}
    \item \textbf{Sinc}:
        \begin{itemize}
            \item Number of hidden layers: 2
            \item Number of neurons in each hidden layer: 200, 200
            \item Activation after each hidden layer: sigmoid, sigmoid
            \item Number of neurons in the output layer: 1
            \item Activation after the output layer: linear
        \end{itemize}
    \item \textbf{Iris}:
        \begin{itemize}
            \item Number of hidden layers: 1
            \item Number of neurons in each hidden layer: 8
            \item Activation after each hidden layer: sigmoid
            \item Number of neurons in the output layer: 3
            \item Activation after the output layer: softmax
        \end{itemize}
    \item \textbf{XOR}:
        \begin{itemize}
            \item Number of hidden layers: 2
            \item Number of neurons in each hidden layer: 16, 8
            \item Activation after each hidden layer: tanh, relu
            \item Number of neurons in the output layer: 1
            \item Activation after the output layer: sigmoid
        \end{itemize}
    \item \textbf{Two Spirals}:
        \begin{itemize}
            \item Number of hidden layers: 4
            \item Number of neurons in each hidden layer: 32, 16, 8, 4
            \item Activation after each hidden layer: tanh, relu, tanh, relu
            \item Number of neurons in the output layer: 1
            \item Activation after the output layer: sigmoid
        \end{itemize}
\end{enumerate}

\begin{table}[]
\centering
\begin{tabular}{|c|c|c|c|c|c|}
\hline
\textbf{Dataset} & \textbf{Runs} & \textbf{RCOND} & \textbf{Iterations} & \textbf{Batch size} & \textbf{Learning Rate} \\ \hline \hline
Boston Housing & 10 & $10^{-15}$ & 2500 & 32 & 0.01 \\ \hline
Sinc & 10 & $10^{-15}$ & 2500 & 32 & 0.01 \\ \hline
Iris & 10 & $10^{-15}$ & 2500 & 32 & 0.01 \\ \hline
XOR & 10 & $10^{-15}$ & 2500 & 32 & 0.01 \\ \hline
Two Spirals & 10 & $10^{-15}$ & 2500 & 32 & 0.005 \\ \hline
MNIST (6-layers) & 1 & $10^{-15}$ & 5000 & 32 & 0.001 \\ \hline
MNIST (8-layers) & 1 & $10^{-15}$ & 5000 & 32 & 0.001 \\ \hline
MNIST (11-layers) & 1 & $10^{-15}$ & 5000 & 32 & 0.0005 \\ \hline
\end{tabular}
\caption{Optimal hyperparameter values for experiments involving the Adam optimizer for Feed-forward Neural Networks. The RCOND value controls the regularization in the pseudoinverse operation, which uses Truncated-SVD. The value was set to numpy's default value.}
\label{tab:hyperparameters}
\end{table}

\subsubsection{Convolution Neural Networks}

Experiments described in section 3 of the main paper involve 3 CNNs. The architecture for each neural network is as follows:

\begin{enumerate}
    \item \textbf{Conv1}:
        \begin{itemize}
            \item Kernel size: 3 x 3
            \item Kernels in each convolution layer: 32
            \item Activations in each convolution layer: sigmoid
        \end{itemize}
    \item \textbf{Conv2}:
        \begin{itemize}
            \item Kernel size: 3 x 3
            \item Kernels in each convolution layer: 32, 64
            \item Activations in each convolution layer: sigmoid, sigmoid
        \end{itemize}
    \item \textbf{Conv3}:
        \begin{itemize}
            \item Kernel size: 3 x 3
            \item Kernels in each convolution layer: 32, 64, 128
            \item Activations in each convolution layer: sigmoid, sigmoid, sigmoid
        \end{itemize}
\end{enumerate}

\begin{table}[h]
\centering
\begin{tabular}{|c|c|c|c|c|c|}
\hline
\textbf{Network} & \textbf{Runs} & \textbf{RCOND} & \textbf{Iterations} & \textbf{Batch size} & \textbf{Learning Rate} \\ \hline \hline
Conv1 & 10 & $10^{-2}$ & 200,000 & 32 & 0.001 \\ \hline
Conv2 & 10 & $10^{-9}$ & 200,000 & 32 & 0.001 \\ \hline
Conv3 & 10 & $10^{-9}$ & 200,000 & 32 & 0.0001 \\ \hline
\end{tabular}
\caption{Optimal hyperparameter values for experiments involving ZORB and the Adam optimizer for Convolution Neural Networks.}
\end{table}
\end{document}